\documentclass[conference]{IEEEtran}
\IEEEoverridecommandlockouts
\usepackage{utfsym}
\usepackage{pifont}
\usepackage{stfloats}
\usepackage{url}
\usepackage{verbatim}
\usepackage{graphicx}
\usepackage{cite}
\usepackage{amsmath,amssymb,amsfonts}
\usepackage{algorithmic}
\usepackage{graphicx}
\usepackage{textcomp}
\usepackage{xcolor}
\usepackage{multirow}
\usepackage{booktabs}
\usepackage{bm}
\usepackage{xcolor}
\def\BibTeX{{\rm B\kern-.05em{\sc i\kern-.025em b}\kern-.08em
    T\kern-.1667em\lower.7ex\hbox{E}\kern-.125emX}}
\begin{document}

\title{ICANet: A Method of Short Video Emotion Recognition Driven by Multimodal Data\\

\thanks{Xuecheng Wu, Mengmeng Tian, and Lanhang Zhai are all with school of Cyber Science and Engineering, Zhengzhou University. \emph{(Corresponding authors: Xuecheng Wu.)}
}
}
\author{\IEEEauthorblockN{Xuecheng Wu}
\IEEEauthorblockA{
\textit{Zhengzhou University}\\
Zhengzhou, Henan 450002, China \\
wuxc@stu.zzu.edu.cn}
\and
\IEEEauthorblockN{Mengmeng Tian}
\IEEEauthorblockA{
\textit{Zhengzhou University}\\
Zhengzhou, Henan  450002, China\\
tmm@stu.zzu.edu.cn}
\and
\IEEEauthorblockN{Lanhang Zhai}
\IEEEauthorblockA{\textit{Zhengzhou University}\\
Zhengzhou, Henan  450002, China \\
zhailhang@163.com}
}

\maketitle

\begin{abstract}
With the fast development of artificial intelligence and short videos, emotion recognition in short videos has become one of the most important research topics in human-computer interaction. At present, most emotion recognition methods still stay in a single modality. However, in daily life, human beings will usually disguise their real emotions, which leads to the problem that the accuracy of single modal emotion recognition is relatively terrible. Moreover, it is not easy to distinguish similar emotions. Therefore, we propose a new approach denoted ICANet to achieve multimodal short video emotion recognition by employing three different modalities of audio, video and optical flow, making up for the lack of a single modality and then improving the accuracy of emotion recognition in short videos. ICANet has a better accuracy of 80.77\% on the IEMOCAP benchmark.
\end{abstract}

\begin{IEEEkeywords}
Multimodal Deep Learning, Emotion Recognition, Attention Mechanism, Human-Computer Interaction.
\end{IEEEkeywords}

\section{Introduction}
People's life always accompanies emotion, which is an essential part of the dynamic mechanism in people's psychological activities. The ``pure cognition" research methods, separated from the emotional factors, can not entirely investigate and simulate human behaviors. Therefore, the fast development of emotion recognition based on neural networks is very significant. At present, emotion recognition methods based on the non-physiological signals mainly include the recognition of facial expression and voice intonation.

However, the faeture information extracted by the single modality emotion recognition method is usually one-sided, resulting in the low accuracy of the recognition. In the contrary, the multimodal fusion method combined with various feature information can more accurately capture and identify human emotions. This paper propose a new method, which combines the three modalities of audio, video and optical flow for multimodal emotion recognition in short videos, making up for the lack of a single modality and improving  the accuracy of emotion recognition in short videos. In this paper, we select the videos, optical flow and LFCC spectrograms as inputs. The three feature extraction networks, I3D (RGB/FLOW) and CA-VGG16, are deployed to model the three modality feature information, respectively, and then the prediction scores corresponding to the three modalities are calculated. The decision level feature fusion is carried out according to a certain modality weight ratio. Finally, the specific feature fusion results are transferred to the SoftMax classifier, and then we get the final emotion recognition results in short videos.

\section{Related Work}
Currently, the research on emotion recognition in short videos is mainly divided into physiological signals based emotion recognition methods and non-physiological signals based emotion recognition methods. The emotion recognition methods based on physiological signals mainly deploy physiological signals such as EEG, EMG, and ECG to predict. Although physiological signals can not be camouflaged and have the capability of getting more objective results, it is relatively difficult to collect these physiological signals. Furthermore, it lacks reasonable evaluation standards, which are unsuitable for practical application. Emotion recognition based on non-physiological signals mainly employs non-physiological signals such as facial expression, voice, and gestures to predict. These signals are relatively easy to collect and do not require much data preprocessing. They can utilize the existing deep learning technologies to predict quickly. The basic idea of multimodal emotion recognition is to predict the emotion category expressed by the detected objects based on the information of text, speech, facial expression and other multimodal inputs. The multimodal emotion recognition method has higher accuracy than previous single modality emotion recognition methods.

\begin{figure}[t]
\setlength{\belowcaptionskip}{-0.4cm}
\centering
\includegraphics[scale=0.345]{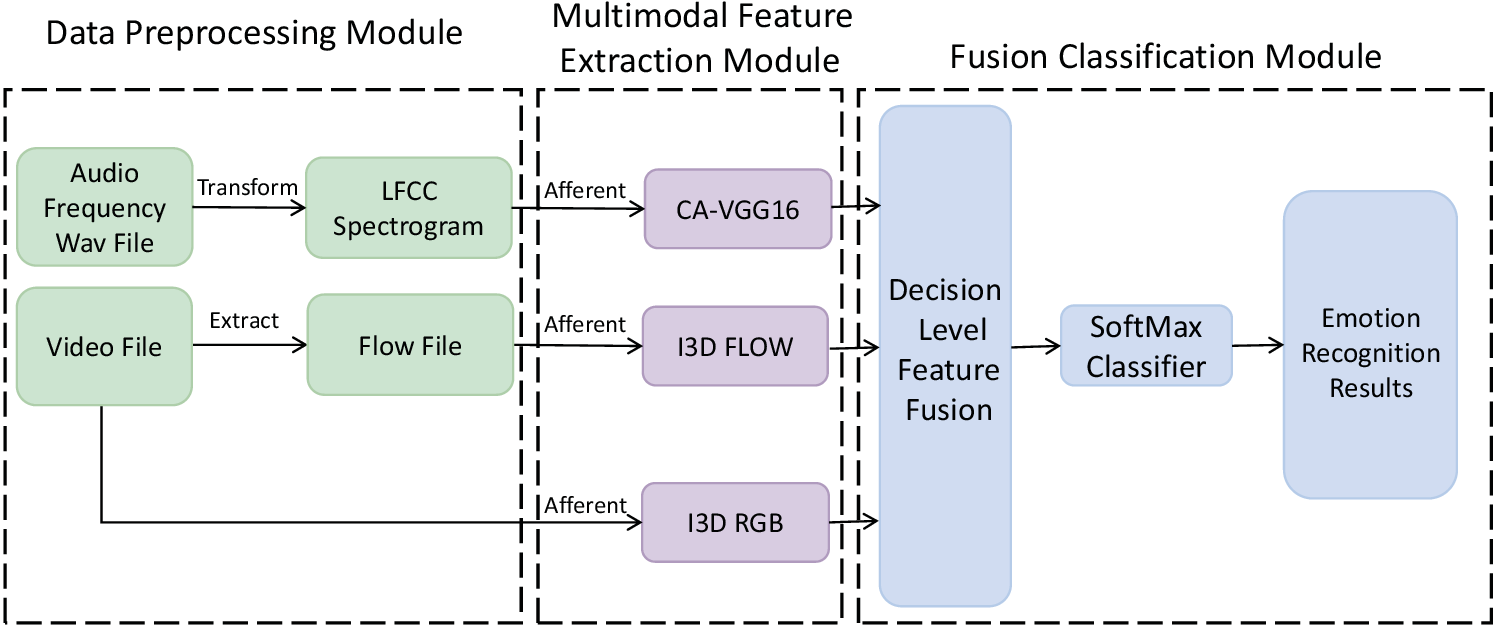}
\caption{The overall illustration of ICANet. It consists of the Data Preprocessing Module, the Multimodal Feature Extraction Module, and the Fusion Classification Module. Specifically, we fuse the three different feature tensors of feature extraction networks in the decision level feature fusion module.}
\label{fig:1}
\end{figure}

\section{Methodology}
The overall structure of ICANet is divided into three parts: data preprocessing module, multimodal feature extraction module and fusion classification module. The overall network structure of ICANet is shown as Fig.~\ref{fig:1}.

\subsection{Data Preprocessing Module}\label{AA}
\subsubsection{LFCC Spectrogram}
In this paper, the audio signal is preprocessed by generating the LFCC spectrograms. LFCC spectrogram is a three-dimensional spectrum representing the voice frequency graph changing with time. Its abscissa is time, its ordinate is frequency, and the coordinate point value is voice data energy. Since the spectrogram deploys a two-dimensional plane to express three-dimensional information, the energy value is expressed by the depth of the colors. Firstly, this paper preprocesses the Wav files of the audio section of the IEMOCAP dataset, using the scipy voice processing tool and python\_speech\_features library to read the speech information and extract the LFCC speech features, and finally converts the ordinary Wav speech signals into LFCC spectrograms.

\subsubsection{Optical Flow Extraction}
Optical Flow is the apparent motion mode of two consecutive interframe images caused by the motion of an object or a camera. It is a 2D vector field, where each vector is a displacement vector, representing the motion of the point from frame $X$ to frame $X+1$. In this paper, we deploy the interface of the optical flow estimation algorithm provided in the visual algorithm library OpenCV, including the sparse flow estimation algorithm cv2.Calcopticalflowpyrlk() and dense flow estimation algorithm cv2.calcOpticalFlowFarneback(). Specifically, we deploy the sparse flow estimation algorithm, which is the Lucas Kanade algorithm \cite{b1}.

\subsection{Multimodal Feature Extraction Module}
\subsubsection{RGB and Flow Feature Extraction Module}
\begin{figure}[t]
\setlength{\belowcaptionskip}{-0.4cm}
\centering
\includegraphics[scale=0.36]{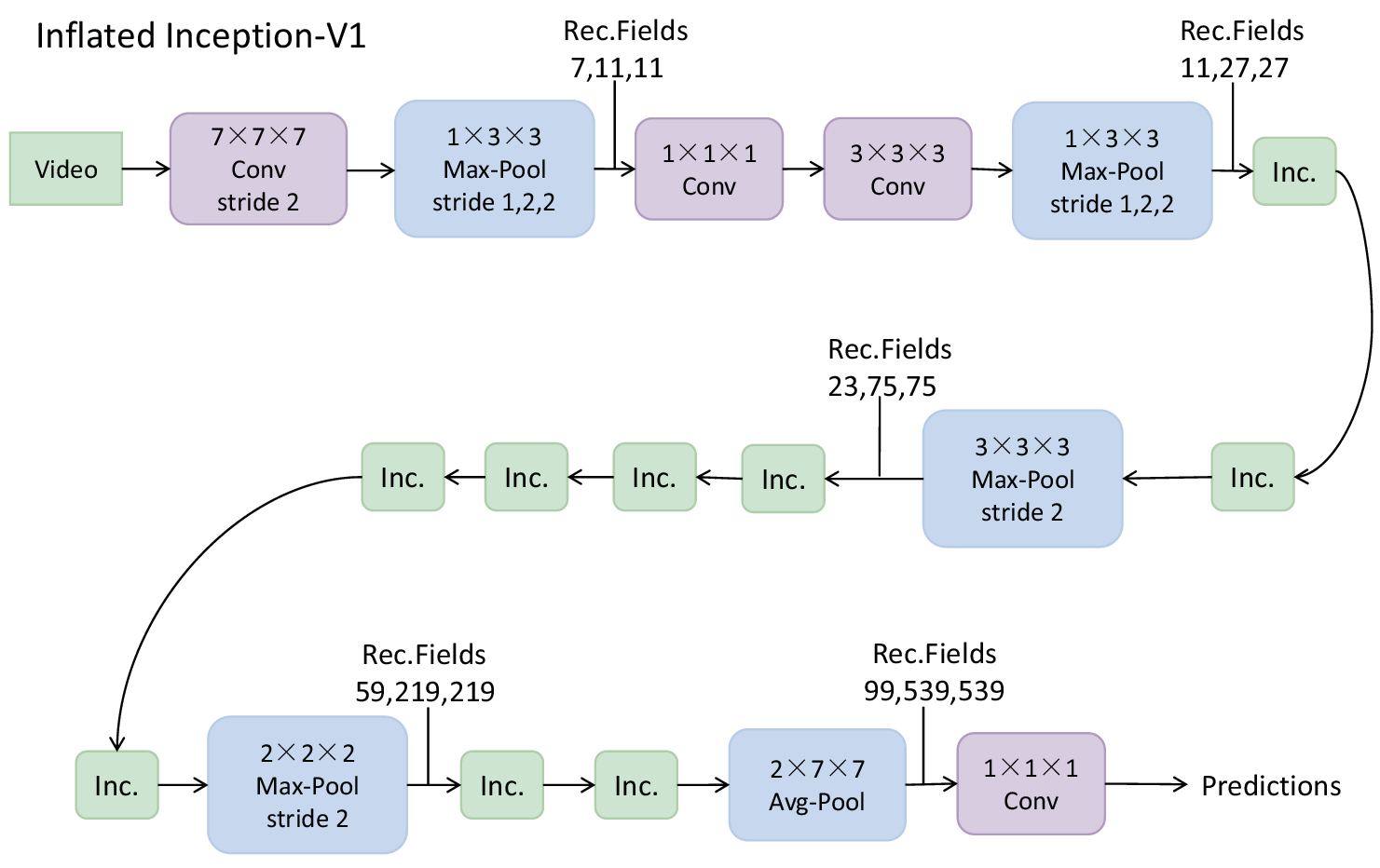}
\caption{The overall network structure of Inflated Inception-V1. Here, ``Rec.Fields" represents the receptive fields for specific feature tensors. }
\label{fig:2}
\end{figure}

\begin{figure}[t]
\setlength{\belowcaptionskip}{-0.4cm}
\centering
\includegraphics[scale=0.46]{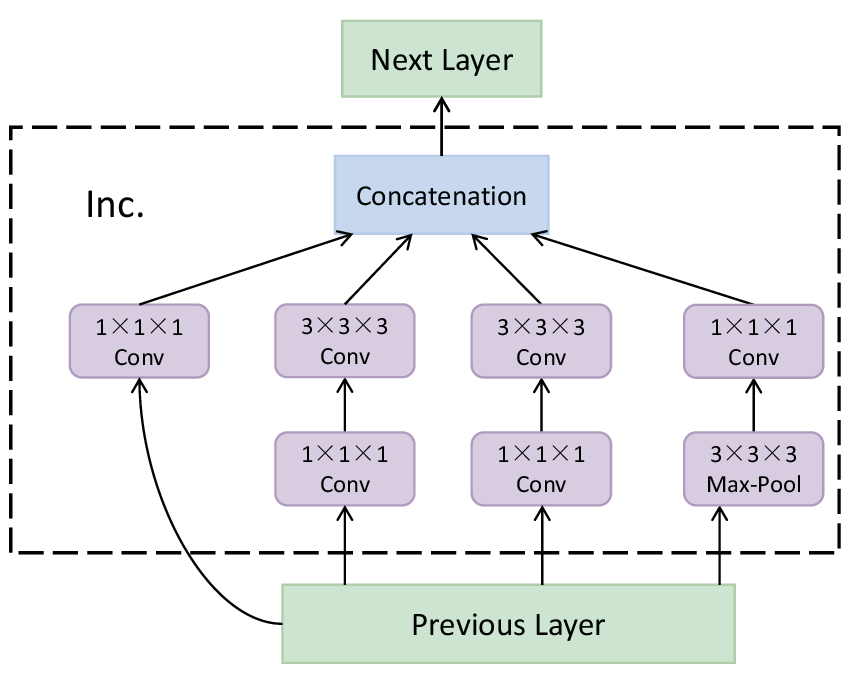}
\caption{The overall illustration of initial space sub module ``Inc.". The strides of convolution and pooling operators are 1, which are not specificed.}
\label{fig:3}
\end{figure}

In this paper, the I3D \cite{b2} is deployed to extract the specific features of the characters in the videos, and the model training method of separating the video stream and the optical flow stream is adopted. I3D (two-stream inflated 3D CNN) \cite{b2} is a video action recognition model proposed by Google DeepMind in 2017, using Inception-V1 as the backbone. Its specific network structure is shown in Fig. \ref{fig:2}. In Inception-V1, the stride of the first convolution transformation layer is 2. In addition to the parallel maximum pooling layer, there are four maximum pooling layers with a stride of $2 \times 2$ and $7 \times 7$. The structure of the initial space sub module ``Inc." is shown as Fig. \ref{fig:3}.

This paper adopts the form of separate training. One I3D network training takes RGB (video) stream as input, and the other takes FLOW (optical flow) stream as input, carrying optimized and smooth stream information. In this paper, the two I3D networks are first trained, and the pre-trained weights are then loaded during model validation. The models are trained on the IEMOCAP  multimodal dataset, and all the video frames have been adjusted to the appropriate size, number and channels before input into the network. Seventy-nine video frames are extracted at uniform time intervals, and the entire video content is captured. Finally, the pre-trained weights of RGB and FLOW modalities are obtained.

\subsubsection{Audio Feature Extraction Module}
\begin{figure}[t]
\setlength{\belowcaptionskip}{-0.4cm}
\centering
\includegraphics[scale=0.4]{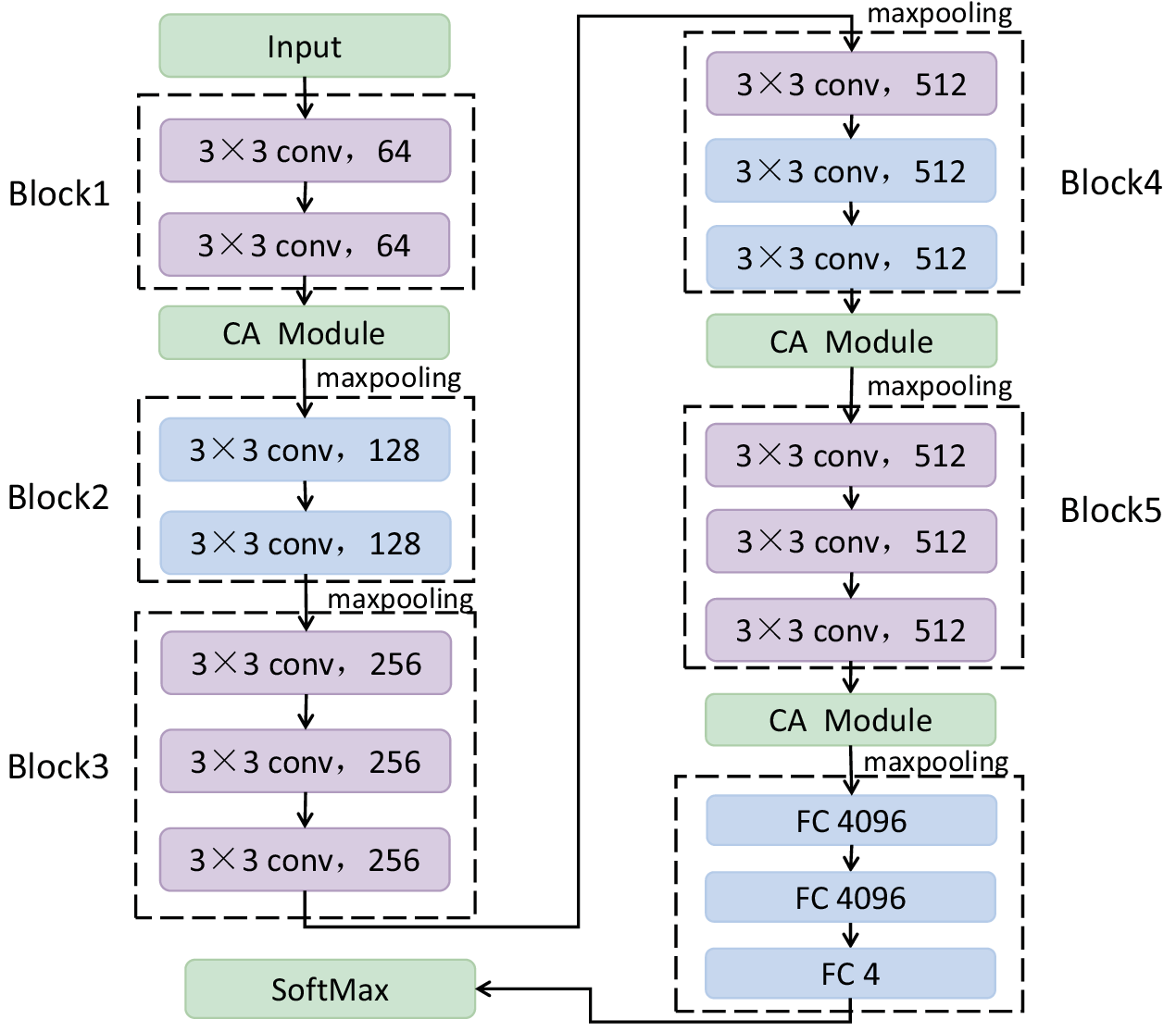}
\caption{The overall network structure of CA-VGG16. Specifically, there are five convolution transformation blocks.}
\label{fig:4}
\end{figure}

This paper first converts audio files into the LFCC spectrograms, and then evolve it into an image classification task. We deploy the independently improved VGG16 denoted as CA-VGG16. We mainly introduce the Coordinate Attention Module (CA module) to improve the performance of VGG16, adding the CA Module after the first, fourth and fifth convolution transformation blocks of VGG16 to improve the model performance. The overall network structure of CA-VGG16 is shown in Fig. \ref{fig:4}.

The Coordinate Attention is a novel attention mechanism proposed by Hou et al. \cite{b3}. The Coordinate Attention innovatively embeds the precise positional information into the inter-channel attention and then improves the capability of the backbone network to capture the detailed object structures, enrich the semantic information of shallow feature maps and obtain the precise positional information of larger areas.

VGG16 \cite{b4} mainly deploys the small convolution filters to build a new convolutional neural network, including the convolution transformation layers, the pooling layers, and the full connection layers.

\textbf{Convolution Transformation Layer}~~~In the VGG16, it proposes to utilize two $3 \times 3$ convolution kernels to replace one specific $5 \times 5$ convolution kernel, and deploy three $3 \times 3$ convolution kernels to replace one $7 \times 7$ convolution kernel. In this approach, we can significantly reduce the parameters of whole network, increase the depth and improve the feature expression ability of the overall network to a certain extent. The calculation formula of the convolution transformation layer are as follows:
\begin{equation}
\mathrm{W}_{\mathrm{n}+1}=\left(\mathrm{W}_{\mathrm{n}}+2 * \mathrm{P}-\mathrm{K}\right) / \mathrm{S}+1
\label{math:1}
\end{equation}
\begin{equation}
\mathrm{H}_{\mathrm{n}+1}=\left(\mathrm{H}_{\mathrm{n}}+2 * \mathrm{P}-\mathrm{K}\right) / \mathrm{S}+1
\label{math:2}
\end{equation}

\textbf{Pooling Layer}~~~VGG16 deploys the maximum pooling layers, followed by a maximum pooling layer after several convolution transformation layers to form a block for compressing the shape of feature maps. The general formula for the pooling layer are shown in the following:
\begin{equation}
\mathrm{W}_{\mathrm{n}+1}=\left(\mathrm{W}_{\mathrm{n}}-\mathrm{K}\right) / \mathrm{S}+1
\label{math:3}
\end{equation}
\begin{equation}
\mathrm{H}_{\mathrm{n}+1}=\left(\mathrm{H}_{\mathrm{n}}-\mathrm{K}\right) / \mathrm{S}+1
\label{math:4}
\end{equation}

From the Eq. \ref{math:1} to the Eq. \ref{math:4}, where $W$ and $H$ represent the width and height, respectively. $P$, $K$, and $S$ represent the filter size, the convolution kernel size, and the stride, respectively.

\textbf{Full Connection Layer}~~~In the CNN, the full connection layer is responsible for the task of ``Classifier", which classifies the feature tensors. VGG16 deploys three full connection layers at the end, and the last full connection layer uses the SoftMax layer for classification.

In this paper, we add the CA modules after the first, fourth and fifth blocks of VGG16, denoted as CA-VGG16-3. At the same time, we also attempt to add the CA modules after each block, denoted as CA-VGG16-5. However, the ablation experimental results show that CA-VGG16-3 performs better than CA-VGG-5. Therefore, we select CA-VGG16-3 as the audio feature extraction network, denoted as CA-VGG16. The detailed network structure of CA-VGG16 is shown in Fig. \ref{fig:4}. 

\subsection{Fusion Classification Module}
The ICANet proposed in this paper adopts the decision level feature fusion module to combine the highest level of pre-trained weights and carry out feature fusion on the prediction score, that is, to take the weights of three modalities, carry out weighted average fusion on them, and then output the final prediction results from a SoftMax classifier.

\section{Experiments}

\subsection{IEMOCAP Dataset Introduction}
In detail, the IEMOCAP \cite{b5} multimodal emotion recognition dataset is deployed to conduct model training and validation. IEMOCAP records the three selected scripts and fictional scene dialogues designed to trigger specific emotions by ten actors. We filter the unbalanced data in the original dataset and finally form an emotion recognition dataset composed of four emotion categories: happiness, sadness, neutrality and anger, with a total number of 5531. The distribution for each category in the IEMOCAP dataset is shown as Tab. \ref{tab:1}.

\begin{table}[]
\centering  
\caption{The distribution for four categories in the IEMOCAP multimodal emotion recognition dataset. }
\resizebox{8.5cm}{!}{
\begin{tabular}{c|c|c|c|c|c}
\toprule
         & \textbf{Happy} & \textbf{Sad}  & \textbf{Neutral} & \textbf{Anger} & \textbf{Total} \\ \midrule 
\midrule
Session1 & 278   & 194  & 384     & 229   & 1085  \\ \midrule 
Session2 & 327   & 197  & 362     & 137   & 1023  \\ \midrule 
Session3 & 286   & 305  & 320     & 240   & 1151  \\ \midrule 
Session4 & 303   & 143  & 258     & 327   & 1031  \\ \midrule 
Session5 & 442   & 245  & 384     & 170   & 1241  \\ \midrule 
\textbf{Total}    & 1636  & 1084 & 1708    & 1103  & \textbf{5531}  \\ \bottomrule
\end{tabular}}
\label{tab:1}
\end{table}

\subsection{Evaluation Indicator}
In the task of multimodal emotion recognition in short videos, we mainly deploy accuracy ($ACC$) as the evaluation indicator. The formula of $ACC$ is shown as Eq. \ref{math:5} below:
\begin{equation}
\mathrm{ACC}=\frac{\mathrm{TP}+\mathrm{TN}}{\mathrm{TP}+\mathrm{TN}+\mathrm{FP}+\mathrm{FN}}
\label{math:5}
\end{equation}

In the abovementioned formula, $TP$ represents a positive sample detected as correct; $TN$ represents a negative sample detected as wrong; $FP$ represents a negative sample detected as a positive sample; $FN$ represents the positive sample detected as a negative sample \cite{b6}.

\subsection{Ablation Studies}
\begin{figure}[t]
\setlength{\belowcaptionskip}{-0.4cm}
\centering
\includegraphics[scale=0.37]{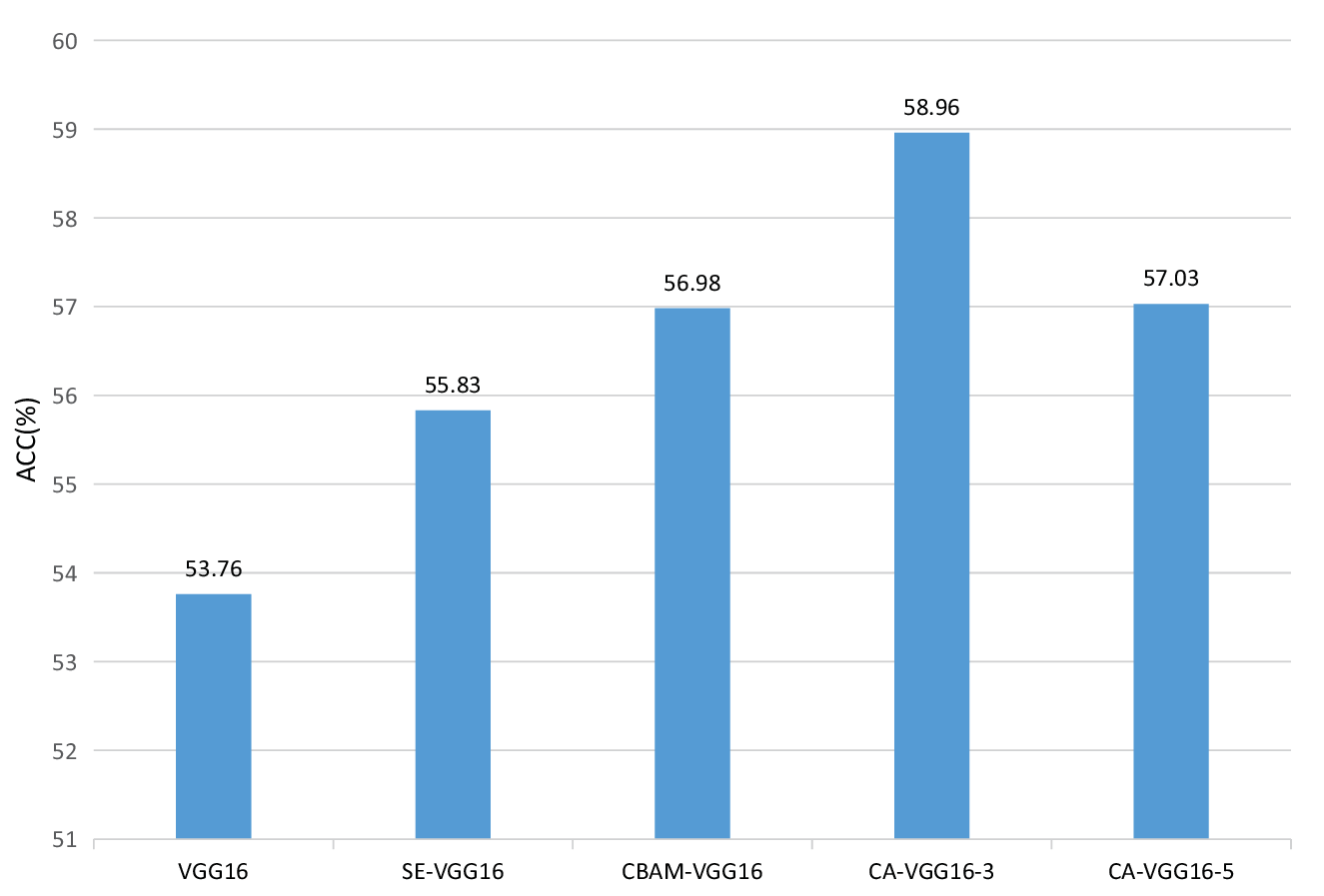}
\caption{The ablation experiments of the specific audio branch of ICANet in terms of ACC(\%) on the IEMOCAP dataset. }
\label{fig:5}
\end{figure}

We conduct four sets of ablation experiments for the audio branch to determine the effectiveness of the CA modules in improving the model performance. The models of these four ablation experiments are original VGG16, SE-VGG16 based on SE Attention, CBAM-VGG16 based on CBAM Attention, CA-VGG16-3 and CA-VGG16-5, respectively. The original VGG16 is set as the control group to determine the effectiveness of the proposed improvement components. The specific ablation results are shown in Fig. \ref{fig:5}.

We can clearly observe that the CA-VGG16-3 achieves the highest accuracy of 58.96\%, which is a practical improvement on the original VGG16. We conclude the reason is that adding the CA modules strengthens the feature extraction capability of the model for LFCC spectrograms and improves the processing of specific features. We have demonstrated that the CA-VGG16 with three CA modules is more feasible and practical in extracting the audio features.

\subsection{Comparison with other SOTA Methods }
To explore the best recognition effect of ICANet, this paper gives different modalities different weight ratios for fusion to obtain the optimal weight ratio. When the weight ratios of RGB, FLOW, and Audio are 1:1:1, 3:2:5, 4:3:3, 5:3:2, respectively, the ACC are respectively 77.50\%, 78.85\%, 79.04\%, 80.77\%, 67.66\%. We can observe that blindly increasing the weight of RGB stream will not achieve better results. On the contrary, moderately increasing the weight of FLOW stream or Audio stream will also improve the overall model performance to a certain extent. 	The reason is that in this approach, the overall network can not achieve good information complementarity and the better effect of multimodal fusion. According to the experimental results, when RGB:FLOW:Audio = 4:2:4, we can get the best model performance. Therefore, we selects 4:2:4 as the optimal weight ratio of three different modalities for later decision level feature fusion.

To further verify the effectiveness of the emotion recognition method in short videos driven by multimodal data proposed in this paper, we compare ICANet with the current mainstream emotion recognition methods. The results of the performance comparison are shown in Tab. \ref{tab:4}. It can be seen from the abovementioned  table that the accuracy of the new emotion recognition method ICANet based on RGB, FLOW and Audio modalities is significantly higher than the above five mainstream emotion recognition baseline methods. Furthermore, the accuracy of ICANet proposed in this paper reaches 80.77\%, exceeding the existing CNN-based state-of-the-art methods by 15.89\%.

The analysis shows that the main reason for the model performance improvement of ICANet is that it integrates the feature information of RGB, FLOW and Audio modalities, realizes the complementarity of feature information, makes up for the deficiency of a single modality, and deploys three sub feature extraction networks with higher accuracy to extract richer emotional feature information. To sum up, the validity of the emotion recognition method in short videos driven by the typical features of multimodal data is verified.

\begin{table}[]
\centering  
\caption{Performance comparison of ICANet and different emotion recognition methods on the IEMOCAP multimodal dataset. We train all the models for the same epoch for a fair comparison.}
\resizebox{8.5cm}{!}{
\begin{tabular}{c|c|c}
\toprule
\textbf{Method }       & \textbf{Modality}       & \textbf{ACC (\%)} \\ \midrule 
\midrule 
I3D           & RGB            & 64.88    \\ \midrule 
I3D           & FLOW           & 60.12    \\ \midrule 
C3D           & RGB            & 61.33    \\ \midrule 
1D Music CNN  & RGB            & 53.04    \\ \midrule 
ResNet50+LSTM & RGB            & 53.37    \\ \midrule 
\textbf{Ours}          & RGB+FLOW+Audio & \textbf{80.77}    \\ \bottomrule
\end{tabular}}
\label{tab:4}
\end{table}

\section{Conclusions}
This paper propose a new emotion recognition method in short videos denoted as ICANet driven by the typical features of multimodal data. The experimental results on the IEMOCAP multimodal emotion recognition dataset show that the accuracy of this new short video emotion recognition can reach 80.77\%. Compared with the current mainstream emotion recognition methods, the recognition accuracy has been significantly improved.

\vspace{12pt}

\end{document}